\documentclass{article}
\usepackage{arxiv}

\usepackage[utf8]{inputenc} 
\usepackage[T1]{fontenc}    
\usepackage{hyperref}       
\usepackage{url}            
\usepackage{booktabs}       
\usepackage{amsfonts}       
\usepackage{nicefrac}       
\usepackage{microtype}      
\usepackage{lipsum}
\usepackage{graphicx}
\usepackage{amsmath}
\usepackage{array}

\makeatletter
\newcommand{\printfnsymbol}[1]{%
  \textsuperscript{\@fnsymbol{#1}}%
}
\makeatother

\title{DGCM-Net: Dense Geometrical Correspondence Matching Network for Incremental Experience-based Robotic Grasping}

\author{
  Timothy Patten\printfnsymbol{1}, Kiru Park\thanks{Equal contribution.}~~, and Markus Vincze \\
  Vision for Robotics Laboratory, Automation and Control Institute, TU Wien, Vienna, Austria\\
  \texttt{\{patten, park, vincze\}@acin.tuwien.ac.at} \\
}

\begin{document}
\maketitle

\begin{abstract}
This article presents a method for grasping novel objects by learning from experience. Successful attempts are remembered and then used to guide future grasps such that more reliable grasping is achieved over time. To transfer the learned experience to unseen objects, we introduce the dense geometric correspondence matching network (DGCM-Net). This applies metric learning to encode objects with similar geometry nearby in feature space. Retrieving relevant experience for an unseen object is thus a nearest neighbour search with the encoded feature maps. DGCM-Net also reconstructs 3D-3D correspondences using the view-dependent normalised object coordinate space to transform grasp configurations from retrieved samples to unseen objects. In comparison to baseline methods, our approach achieves an equivalent grasp success rate. However, the baselines are significantly improved when fusing the knowledge from experience with their grasp proposal strategy. Offline experiments with a grasping dataset highlight the capability to transfer grasps to new instances as well as to improve success rate over time from increasing experience. Lastly, by learning task-relevant grasps, our approach can prioritise grasp configurations that enable the functional use of objects.
\end{abstract}

\section{Introduction}

Grasping is an essential capability for robots in a large variety of fields, from warehouse operations to industrial assembly lines, applications in agriculture and many domestic service tasks. Grasping leads to the subsequent manipulation of objects, which is the most direct way for robots to interact with the world. Especially in human environments, where many man-made objects are designed to be handled by people, grasping with a robotic gripper or hand is necessary.

A popular approach for robot grasping is to exploit known objects~\cite{Klank2009, Srinivasa2010, Chitta2012, Tremblay2018_DeepOP, Wang2019_DenseFusion}. However, this can only be applied to a given set of objects and thus does not generalise to new objects, which reduces the usability for real-world operation. It is possible to grasp unknown objects by learning classifiers, predictive or generative models~\cite{Saxena2008, Jiang2011, Fischinger2015_HAF, Lenz2015, Redmon2015, Pinto2016, Kumra2017, Wang2017, Morrison2018} but large amounts of labelled data are required and this is time consuming when it is annotated by hand. The effort for annotating data is eased by restricting it to 2D grasp poses (i.e., assuming top-down grasps), but this limits the grasp configurations that are able to be applied. Training data is often generated without human annotation offline with collections of object models~\cite{Mahler2016_DexNet1, Mahler2017_DexNet2} or online with real-world robot trials. However, many hundreds or thousands of hours are required to generate a sufficient amount of data~\cite{Pinto2016, Jang2017}. Learning end-to-end strategies for grasping with reinforcement learning~\cite{Boularias2015, Levine2018, Kalashnikov2018, Zeng2018} also suffers from substantial training time, with some work reporting training times in the order of months~\cite{Levine2018}. While the burden of learning time can be alleviated by leveraging physics-enabled simulation environments, e.g.,~\cite{James2017, Fang2018, Iqbal2019, James2019}, this introduces the challenge of transferring from simulation to the real world.

An alternative approach is to transfer grasps for known objects to familiar objects. This makes the assumption that when a new object is similar to another object for which a grasp is known, then the new object is likely to be successfully grasped in a similar way~\cite{Bohg2014}. Prior work on experience-based grasping build a database of sensory observations with associated grasp information such as a pose or contact points. The experience is accumulated by trial and error with a robot platform~\cite{Morales2004, Herzog2012, Detry2013b}, kinesthetically taught~\cite{Kroemer2012, Detry2013, Kopicki2016} or inferred by directly observing human behaviour~\cite{Liu2019}. Grasping an unseen object requires a strategy to map the current observation to the samples in the database and execute (or extrapolate from) the most similar experience. This is typically done using global shape~\cite{Morales2004, Bohg2009, Kopicki2016}, local descriptors~\cite{Liu2019} or object regions~\cite{Kroemer2012, Herzog2012, Detry2012, Detry2013, Detry2013b}. In contrast to end-to-end learning approaches, experience-based grasping has the potential to learn from very few exemplars. Only few methods have been demonstrated in an end-to-end pipeline with a robotic platform and currently they rely on hand-crafted features for retrieving similar experiences and for transferring grasps to new objects.

In this work we present a new method for incremental grasp learning from experience. The key to our approach is to apply dense geometrical correspondence matching. Familiar objects are identified through global geometric encoding and associated grasps are transferred through local correspondence matching. We introduce the dense geometrical correspondence matching network (DGCM-Net) that uses metric learning to encode the global geometry of objects in depth images such that similar geometries are represented nearby in feature space to allow accurate retrieval of experience. DGCM-Net additionally reconstructs dense geometrical correspondences between pairs of depth images using a variant of normalised object coordinate (NOC) values. These values are used to compute the rigid transformation between the local region around the grasp of a stored experience and the corresponding region on an object in a new scene. Precise 3D object models are not assumed, thus we define the view-dependent normalised object coordinates (VD-NOC) to extend NOC values to single views.

DGCM-Net is applied in an incremental grasp learning pipeline, in which a robot self-supervises grasp learning from its own experience. We show that a robot learns to repeatably grasp the same object after one or two successful experiences and also to grasp novel objects that have comparable geometry to a known experience. As an extension, we show that the predictions from DGCM-Net improve the performance of baseline grasping methods by combining their quality measures with our experience-based measure. The incremental learning pipeline is also flexible in that grasp success is not the only measure to constitute experience. Specific positions or configurations of grasps can be preferred and therefore used in future situations. In particular, semantic grasps, such as grasping the handle of a mug, are prioritised as they are more relevant for the subsequent manipulation of the object~\cite{Song2010, Dang2012, Fang2018, Antonova2018}. As a result, task-oriented grasps are quickly learned, allowing a robot to perform meaningful actions with objects.

Studies with a dataset showcase the ability of the presented grasp prediction method to transfer between objects. The results show strong within class transfer as well as between class generalisation. In addition, our analysis confirms the intuition that the quality of grasp prediction improves with increasing experience. Real-world experiments with a mobile manipulator are performed to compare our grasping strategy against various other approaches. The experiments show that we achieve a comparable grasp success rate with the baselines and improve the baseline when integrating our predictions to achieve superior performance overall. Demonstrations of the full system show the continuous learning capability for completely novel objects from classes never before seen. Finally, the usability of our approach for semantic or task-oriented grasping is illustrated to grasp objects with handles.

In summary, this article makes the following contributions:
\begin{itemize}
    \item The dense geometrical correspondence matching network to encode object geometry for nearest neighbour retrieval and to densely reconstruct 3D-3D correspondences in order to transfer grasps from stored experiences to unseen objects.
    \item An experience-based 6D grasp learning pipeline that incrementally grows a database of exemplars to guide grasp selection for the same object or novel unseen objects.
    \item Offline experiments with a new annotated dataset showing the capability of DGCM-Net to generalise grasps to unseen objects as well as to steadily improve over time with increasing accumulation of data.
    \item Online grasping experiments showing that the grasp success of our approach is competitive with common baselines and improves the baselines when combining their predictions with our experience-based grasp predictions.
    \item Demonstrations showing the extension of our method for semantic grasping by guiding grasp selection to the parts of objects that are relevant to the object's functional use.
\end{itemize}

The remainder of the paper is organised as follows. Section~\ref{sec:related_work} discusses related work. In Section~\ref{sec:methods}, we present the dense geometrical correspondence matching network and describe the incremental grasp learning pipeline. The results of the offline and robotic experiments are reported in Section~\ref{sec:ablation_studies} and Section~\ref{sec:experiments}. Finally, Section~\ref{sec:conclusion} concludes and discusses future work.

\section{Related Work}\label{sec:related_work}

The significant amount of attention given to robotic grasping has resulted in a large number and high diversity of techniques. A common strategy uses known object instances, which are provided as CAD models or are captured by a modelling process (e.g.,~\cite{Prankl2015_RTMT, Wang2019_Modelling}). Given a known grasp configuration for an object in its local coordinate system, the task of grasping is simplified to recognising and estimating the pose of the object such that the grasp pose is transformed into the new scene. Traditional methods identify hand-crafted features to localise an object model within a scene~\cite{Klank2009, Srinivasa2010, Chitta2012} but more recently advances for pose estimation have been made by the application of deep learning~\cite{Xiang2017_PoseCNN, Park2019_Pix2Pose, Zakharov2019_DPOD, Li2019_CDPN} and grasping pipelines achieve high success rate~\cite{Tremblay2018_DeepOP, Wang2019_DenseFusion}. The main limitation of this direction of research, however, is the closed-world assumption. The approach is restricted to only the objects for which a model is provided and thus cannot generalise to unknown objects.

To address the problem of grasping unknown objects, local geometry can serve as a strong cue. For example, fitting primitives and estimating grasps based on the geometrical structure of the primitives~\cite{Rusu2009} or fitting superquadratics and synthesising grasp poses at the points of minimum curvature~\cite{Makhal2018} have been shown to work in certain cases. More often though, unknown object grasping is addressed by learning from data~\cite{Bohg2014}. Along this line, methods predict the success of a proposed grasp by training a traditional classifier~\cite{Jiang2011, Fischinger2015_HAF} or deep neural network~\cite{Saxena2008, Lenz2015, Redmon2015, Pinto2016, Kumra2017, Wang2017}. Alternatively, grasp simulation or analytical grasp metrics are computed for objects in model databases to generate training data~\cite{Johns2016, Mahler2016_DexNet1, Mahler2017_DexNet2, tenPas2017_GPD, Liang2019_PointNetGPD, Mousavian2019_GraspNet, Cai2019}. The task is then to learn a model that can predict the value of the grasp metric given a proposal and then select the grasp that is most likely to succeed. There is also work that avoids the sampling and scoring procedure by directly predicting a grasp pose with a quality measure~\cite{Morrison2018}. The generative method has proven to be computationally superior and sufficiently fast to be integrated in a closed-loop system. While the work for unknown object grasping has made considerable achievements, they are limited by the diversity of the training data. Out of distribution objects may not receive accurate grasp quality predictions and may fail. Thus, it is necessary to continuously learn and add new examples to the training set. Unfortunately, the deep neural networks that are applied do not have the capacity to be updated online. Our approach, on the other hand, does not need to retrain for grasp prediction. By abstracting the learning component to correspondence matching, we simply add experience to a database and use the network to predict the closest matches for grasp transfer.

Another approach to grasping is to leverage real robot experience and learn end-to-end strategies. One direction is to employ reinforcement learning~\cite{Boularias2015, Levine2018, Kalashnikov2018, Zeng2018}. The advantage of an end-to-end approach is that complete grasping policies can be learned directly from visual input, which removes the need for a dedicated perception pipeline with an additional motion planner for execution. A disadvantage, however, is that the policies can only be applied to scenarios that are perceptually similar, and thus generalisation to novel environments is limited. Unsupervised methods, such as~\cite{Jang2018_Grasp2Vec}, better generalise to unseen scenarios and objects. They are more general to the task and less sensitive to the training scenes. These methods learn an embedding that can be used to retrieve manipulation policies for online execution. Despite these advances, the major drawback of both self- and unsupervised learning is that many attempts are needed for training. Our approach, in contrast, only needs a handful of experiences to reliably repeat past successes. Although physics simulation is now a popular alternative for training learning algorithms, the transfer from simulation to the real world requires additional attention~\cite{James2017, Fang2018, Iqbal2019, James2019}.

Experience-based grasping is much more efficient than reinforcement learning methods since far fewer examples are needed to learn grasps. The common approach is to accumulate samples of past success or failure to guide the grasp selection in new scenarios, under the assumption that objects with similar shape (or appearance) can be grasped in a similar way. Some work define global shape descriptors and train a discriminative classifier to identify the similarity between object shapes to transfer grasps to familiar objects~\cite{Morales2004, Bohg2009, Kopicki2016}. Other work leverage local features descriptors to identify the relevant local regions associated with contact points to transfer grasps between objects within the same class~\cite{Liu2019}. Another approach is to analyse object regions and to maintain a library of prototypical grasps for recurring object parts. This is accomplished by measuring the similarity between regions on the surface of objects such as with height maps~\cite{Herzog2012} or by surface distributions or densities~\cite{Kroemer2012, Detry2012, Detry2013, Detry2013b}. A major assumption is that the observed parts are equivalent, which means grasp transfer is the application of a transformation from the prototype to the scene. They do not deal with the possibility of scale change or deformation. Such geometry variation would have to be stored as a new experience. Our approach deals with this challenge by aligning shapes through dense 3D-3D correspondences and thus also modifies the shape of the grasp to fit the new geometry. Another drawback of prior work is that they use hand-crafted features to encode shape information. We instead generate descriptive features through metric learning, which has been shown to be powerful for similar geometric matching tasks~\cite{Zeng2016_3DMatch}.

Our approach for grasping relies on first finding the nearest observation in a database and second predicting dense geometric correspondences to transform a successful grasp pose to a new observation. Retrieving similar samples has been addressed using learned feature descriptors from RGB-D images~\cite{Wohlhart2015_learning, Balntas2017_pose_guided, Park2019_MTTM}. These employ the triplet loss to train a network to produce smaller feature distances for pairs of images with similar view points while producing larger feature distances for pairs of images with different view points or those that contain different object classes. However, retrieving a similar viewpoint is insufficient when samples do not cover the entire object pose or when target objects are not constrained to a fixed set of a classes. This motivates our method that predicts geometric correspondences between images to match local areas regardless of different scales or detailed shapes. A simple approach is to encode and match local feature descriptors that represent local shapes in 3D point clouds or depth images~\cite{Zeng2016_3DMatch, Zeng2018_matching}. Leveraging this idea, determining pixel-wise correspondences has been demonstrated in a pipeline to predict local correspondences or key-points while considering global contexts of objects~\cite{Florence2018_DenseObjectNets, Manuelli2019_kPAM}. The task of this line of research is to find the corresponding points between an input and a known reference object, therefore, they are not applicable when the reference image has to be selected from various objects and view points based only on global shape.

In order to make this extension, we employ the normalised object coordinate space that has been used to estimate the 6D pose of instances~\cite{Park2019_Pix2Pose, Li2019_CDPN} and classes~\cite{Wang2019_NOCS}. Since NOC values represents coordinate values in the object's local frame and correspondences between the object model and the scene, predicting NOC values is sufficient for computing the transformation between local points from one observation to another. However, it is difficult to define NOC values without knowing the full 3D shape of an object or a common representation for a class. For our work, it is necessary to predict dense correspondences between pairs of images that have similar geometry. To that end, we represent NOC values in the reference frame of the camera view point instead of the object. This alteration to the NOC representation is referred to as the view-dependent normalised object coordinate space. The prediction of VD-NOC values is used to compute the transformation of local areas that are relevant to grasps in order to transform stored grasp poses to the object in the input image. 


\section{Method}\label{sec:methods}

This section describes our methodology for incremental experience-based grasp learning. We begin with an overview of the framework. We then describe the dense geometrical correspondence matching network for retrieving experience samples and for generating dense 3D-3D correspondences. Lastly, we outline how grasps are transferred between local regions using the predicted correspondences.

\subsection{Incremental Grasp Learning Framework}

\begin{figure}[t!]
\begin{center}
\includegraphics[width=0.95\textwidth]{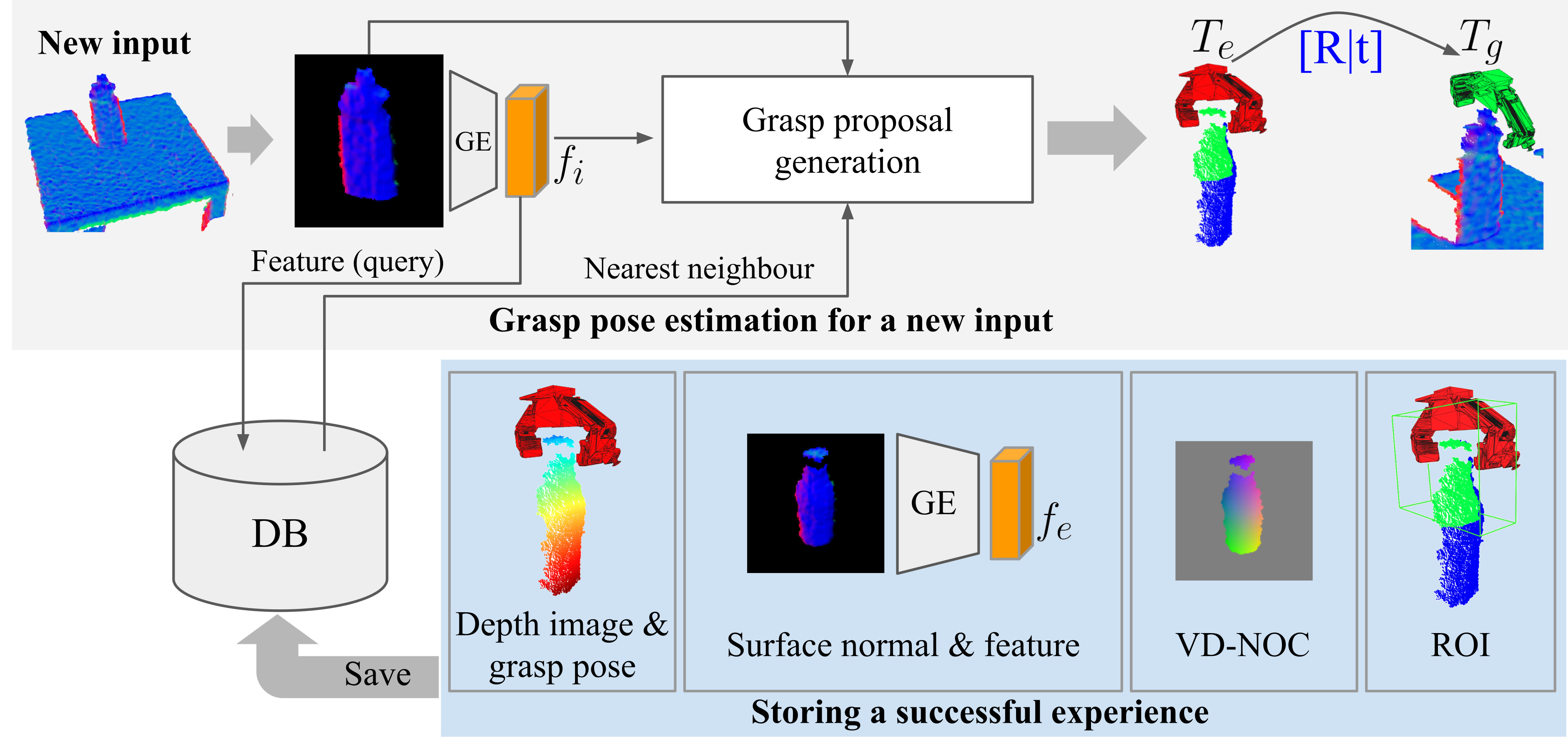}
\end{center}
\caption{Overview of storing and retrieving experience with the incremental grasp learning framework.}\label{fig:est_process}
\end{figure}

The main components of the incremental experience-based grasp learning framework are shown in Figure~\ref{fig:est_process}. The input is a depth image $D_i \in \mathbb{R}^{W \times H}$ and a segmentation mask $M_i \in \mathbb{R}^{W \times H}$ that has entries 1 for pixels belonging to the target object and 0 otherwise. The goal of the framework is to generate a pose for the gripper that will result in a successful grasp. This is represented as a rigid transformation $T \in SE(3)$ of the gripper pose in the camera coordinate frame.

The first step is to match the target object to samples stored in an experience database $\mathcal{E}$. Matching is done using the global geometric encoding from DGCM-Net, where the feature map $f_i$ of the input image is compared to the feature maps of the database samples. Feature maps are the output of a geometry encoder that takes as input a surface normal image derived from the initial depth image. The set of samples with high matching score are used to propose a candidate grasp. For each database match in $\mathcal{E}$, the output of the VD-NOC encoder $c_e$ and the geometry feature encoding $f_e$ as well as $f_i$ from the input are passed to the decoder of DGCM-Net to reconstruct the VD-NOC values $V_i \in \mathbb{R}^{W \times H \times 3}$. This represents a dense mapping between the pixels of the sample and the input and thus a transformation of the points in the 3D coordinates can be computed. Each experience has an associated grasp pose, therefore, the transformation between the images is applied to transform the experience grasp to the target object. Sensitivity to the difference in geometry between the input and sample is reduced by confining the alignment to the region around the grasp pose. The region of interest (ROI) on the sample $\mathcal{R}_e$ is determined from the overlap of the gripper with the 3D coordinates of the segmented object in the depth image. The corresponding ROI on the target object $\mathcal{R}_i$ is derived through the matches between the VD-NOC values. The ROIs are aligned by finding the optimal rotation and translation. The outcome is a proposal for a full 6D grasp pose for the target object.

Incremental learning operates by executing a selected grasp proposal and updating the database online with a new exemplar if the grasp is successful. Specifically, the depth image, surface normal image, VD-NOC values, ROI and transformation of the grasp pose are stored. Unsuccessful grasp attempts do not provide any information for replicating past experience, therefore, no data is stored for failed grasps. As more experience is accumulated, the likelihood of finding a nearby match for a new input increases. The method is not restricted to only finding samples of exact object instances, but can match to new or unseen objects if they have geometry resembling those from experience.

\subsection{Dense Geometrical Correspondence Matching}\label{sec:dense_geo_corr_matching}

\subsubsection{View-dependent Normalised Object Coordinate Space}

Predicting dense correspondences between two depth images (i.e., the depth image of the object to grasp and an experience in the database) is done by predicting a variant of NOC values. The traditional NOC values represent the correspondence between the target object and another one in the target object's local frame. Typically this has been applied for object pose estimation where the target object is a reference model and the other object is an observation of the reference model in a scene.

To apply the same methodology without object models, we introduce the view-dependent normalised object coordinate values. The depth images for a reference and an input are converted to surface normal images. The VD-NOC values for the input $V_i$ are computed using the 3D coordinates of each pixel $I_i^{3D}$ from the input segmentation mask in the camera coordinate frame. Normalisation is performed by setting the origin to the mean coordinate between the maximum and minimum values of $I_i^{3D}$ according to, 
\begin{equation}\label{eq:vd-noc}
V_i = \frac{I_i^{3D} -  \overline{I}_i^{3D}}{\max|I_i^{3D} -  \overline{I}_i^{3D}|}, \hspace{1ex} \textrm{where}  \hspace{1ex} \bar{I}_i^{3D}=\frac{\max(I_i^{3D}) + \min(I_i^{3D})}{2}.
\end{equation}
Normalisation is performed separately for each dimension resulting in different normalisation factors for each axis. The direction of the z-axis is flipped to produce positive values for points that are nearer.

For grasping, the VD-NOC values are used to estimate the similarity between points on the target object in the input image and the points on the object in the experience database. A smaller distance between values in the VD-NOC values represents closer geometrical correspondence. These can be used to estimate the transformation of a set of points in the grasp pose ROI in order to transfer the grasp experience to the target object. 

\subsubsection{DGCM-Net Architecture}

\begin{figure}[t!]
\begin{center}
\includegraphics[width=0.95\textwidth]{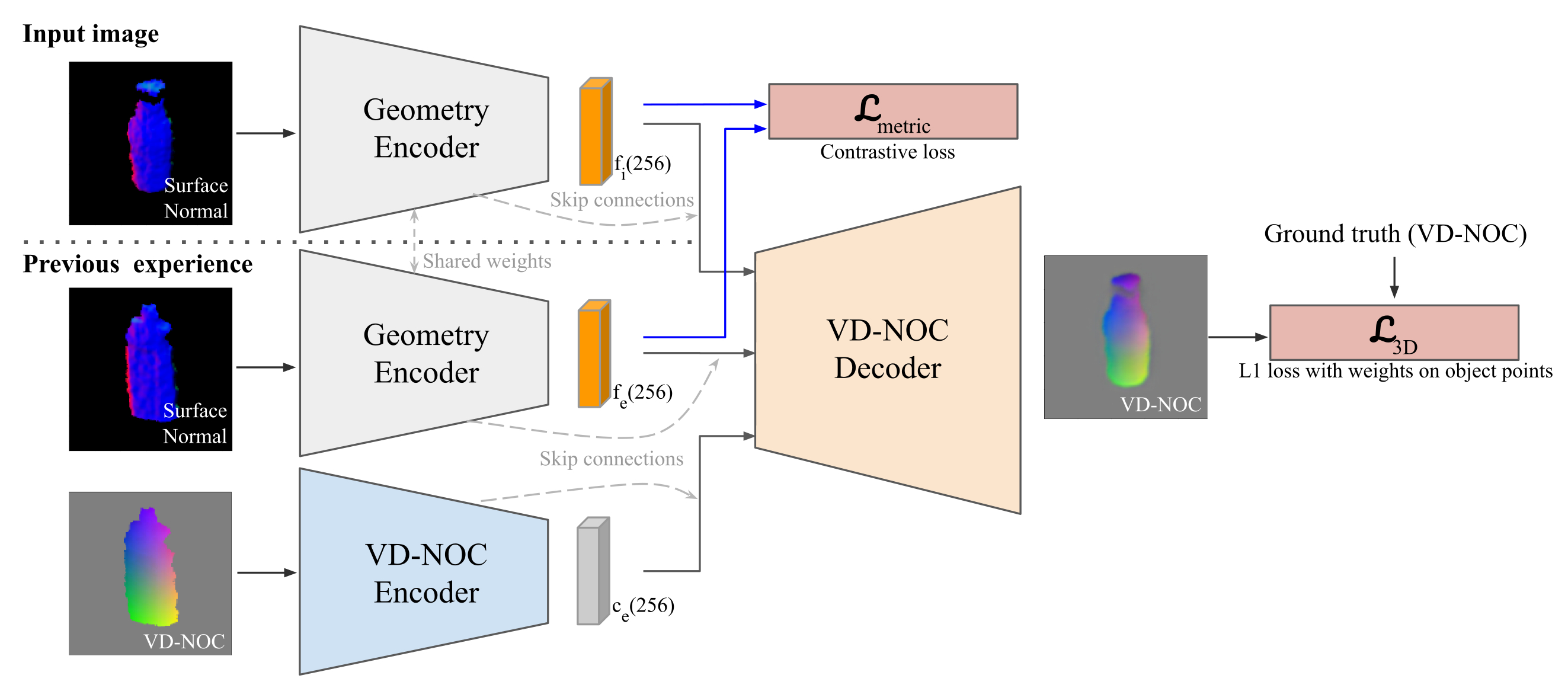}
\end{center}
\caption{Overview of the DGCM-Net architecture and training objectives.}\label{fig:network_arch}
\end{figure}

An overview of the dense geometric correspondence network is shown in Figure~\ref{fig:network_arch}. DGCM-Net consists of a geometry encoder, VD-NOC encoder and VD-NOC decoder. The purpose of the geometry encoder is to learn a representation that places images with similar geometry closer in feature space than images with dissimilar geometry. The purpose of the VD-NOC encoder-decoder is to reconstruct the VD-NOC values between a pair of images.

The input to the geometry encoder is a cropped surface normal image derived from the input depth image and segmentation mask. The cropped image is created from a 2D bounding box that is centred at the 2D projected point of the segmentation mask's centroid. The height and width of the image are adjusted to correspond to 30cm spatial size in 3D space. The cropped image is then resized to 128x128. The first three blocks of the Resnet-50~\cite{He2016_resnet} architecture is employed and initialised with the pre-trained weights using the ImageNet dataset~\cite{Deng2009_ImageNet}. The output of the third block is passed to three convolution layers, kernel sizes = [3, 3, 2] and filter sizes = [256, 256, 128] with strides 2 for all, and two fully connected layers with 256 outputs. The \textit{LeakyReLU} activation is applied to every layer output except the last layer that uses the $\tanh$ as an activation to transform feature descriptors to 256 dimensions.

The input to the VD-NOC encoder is a cropped VD-NOC image. The input is passed to five convolution layers, kernel sizes = [5, 3, 3, 3, 3] and filter sizes = [128, 256, 256, 256, 256] with strides 2 for all, and one fully connected layer with 256 outputs. The activation of each layer is the same for the geometry encoder. The VD-NOC decoder reconstructs the VD-NOC values for the input image with respect to its camera frame. The input to the decoder is the concatenated features from both geometry encodings of the images and the output of the VD-NOC encoder for the reference. Skip connections~\cite{Ronneberger2015_unet} are added by concatenating one half of the output channels of each intermediate layer of the encoders with corresponding layers in the decoder. This helps to predict fine details in local areas. The decoder ends with a fully connected layer with 2048 outputs followed by five blocks of deconvolution and convolution layers. The output of the last convolution layer is the same size as the input image with three channels that represent the x, y and z of the VD-NOC values.

\subsubsection{Training Objective}

DGCM-Net has two tasks and therefore consists of two objectives in the training process. The first is the metric learning of feature descriptors to perform matching and the second is for reconstructing the VD-NOC values of an input image. For metric learning, the contrastive loss~\cite{Hadsell2006_Contrasiveloss} is employed to minimise the Euclidean distance between features of similar geometry (positive pair) while increasing the distance for a pair of different geometry (negative pair) as formulated by,
\begin{equation} \label{eq:metric_loss}    
\mathcal{L}_\textrm{metric} = {{1} \over{N}} \sum^{N}_{i=1} (1-\omega_i)d_i^2+\omega_i\max(10-d_i,0)^2,
\end{equation}
where $\omega$ denotes labels for pairs that are set to 0 for positive pairs and 1 for negative pairs. $d$ denotes the Euclidean distance between encoded feature vectors ($f_i, f_e \in \mathbb{R}^{256}$) of the target and experience images from the geometry encoder. The loss is computed for a mini-batch that consists of $N$ pairs of training images.

For the reconstruction of VD-NOC values, the standard L1 loss is applied for each pixel $p$. Since background pixels are masked out, their values are easy to predict. Hence, the loss values for pixels on the object masks $M_i \in \mathbb{R}^{W\times H}$ are weighted by a factor of 3 to more precisely predict the values of pixels in the object masks~\cite{Park2019_Pix2Pose}. The reconstruction loss is thus given by,

\begin{equation} \label{eq:recont_loss}    
\mathcal{L}_\textrm{3D} = \frac{1}{N\times W \times H} \sum^N_{i=1}{\left[ 3\sum_{p \in M_i}||V_\textrm{i}^p-V_\textrm{gt}^p||_1 + \sum_{p \notin M_i} ||V_\textrm{i}^p-V_\textrm{gt}^p||_1 \right]}.
\end{equation} 
The reconstruction loss is computed only if the pair of samples is positive. Finally, the objective of the training is the weighted sum of two loss functions,
\begin{equation} \label{eq:all_loss}    
\mathcal{L} = \mathcal{L}_\textrm{metric}+\lambda\mathcal{L}_\textrm{3D},
\end{equation} 
where $\lambda$ is a weight balancing the two objectives. We set $\lambda$ to 1 in our experiments.

\subsubsection{Training using Synthetic Images}

\begin{figure}[t!]
\begin{center}
\includegraphics[width=0.95\textwidth]{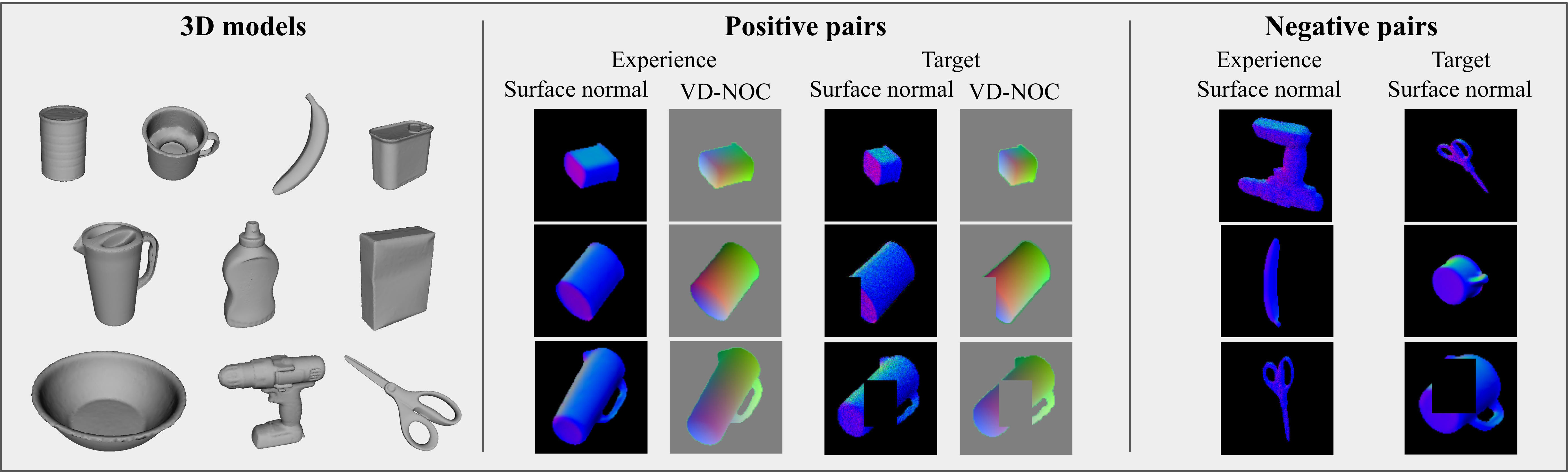}
\end{center}
\caption{Left: Object models from the YCB dataset used to train DGCM-Net. Middle: Examples of positive pairs with online augmentation, Gaussian noise and partial occlusion. Right: Examples of negative pairs. The pairs are used to train feature vectors to have smaller distance for similar geometries.}\label{fig:training_obj_network}
\end{figure}

Synthetic depth images are created to train the network. 3D models are sampled such that no two models are the same even after a scale change\footnote{The set of 3D models only contains one box because any other box can be constructed just by manipulating the scale in the different dimensions.}. Objects are selected from the YCB object and model set~\cite{Calli2017} and listed in Figure~\ref{fig:training_obj_network}. Depth images are rendered in OpenGL\footnote{https://www.opengl.org} for each object model by uniformly sampling a pose and randomly selecting scale factors for each axis. Five scenes are rendered with different scales for each sampled object pose. To avoid ambiguous views of symmetric objects, view angles are limited between 0 and 45 degrees on each axis. For cylindrical objects, no variation around the rotational axis is applied. Parameters used in the generation process are summarised in Table~\ref{tab:train_data_param}. The result for every training sample is a depth image, VD-NOC image, annotated pose, annotated scale factors for each dimension and a look-up table of visible vertices. Approximately 166k images are created and used for training.

\begingroup
\renewcommand{\arraystretch}{1.4}
\begin{table}[t!]
\begin{centering}
\begin{tabular}{p{2.5cm}|c|c|c|c}
  \toprule
  Stage & \multicolumn{2}{c|}{Data generation} &  \multicolumn{2}{c}{Online augmentation}\\
  \hline
  Parameters &Scale (each axis) &  Distance to camera  &  Frac. of occluded region & Gaussian noise \\
 \hline
   Range &$\mathcal{U}$(0.8, 1.5) & $\mathcal{U}$(1m, 1.7m)&$\mathcal{U}$(0.0, 0.25)&$\mathcal{N}$($\mu$=0,$\sigma$=0.01)\\
  \bottomrule
\end{tabular}
\caption{Overview of the parameters used to generate the training data and for the online data augmentation.}\label{tab:train_data_param}
\end{centering}
\end{table}
\endgroup

Metric learning requires positive and negative pairs. Positive pairs are obtained from samples of the same object in different poses when a pair of images share more than half of the visible vertices. Negative pairs are obtained from different objects or different poses of the same object when images share less than half of the visible vertices. Examples of training samples of both types are given in Figure \ref{fig:training_obj_network} (middle and right). For positive pairs, the target VD-NOC values (i.e., the ground truth value) is computed using the relative pose of the object, which is known for the training samples. Thus, the VD-NOV values that are defined in the camera frame of the first element of the pair are transformed to the camera frame of the second element. For our grasping framework, this amounts to transforming the VD-NOC values from the object in the input image to the object in the experience database.

Further augmentation is applied to the image samples to improve robustness against occlusion and noise. Occlusion is simulated by setting a partial area in the surface normal image and the corresponding entries in the VD-NOC values to zero (i.e., the value for the background). This enables the network to learn features that still return good matches between an input and samples in the database even when one is occluded. Gaussian noise is also applied to both images to cope with the expected noise from real sensors. Figure~\ref{fig:training_obj_network} (middle and right) presents examples after applying the augmentation. More details about the parameters used for augmentation are provided in Table~\ref{tab:train_data_param}.

We train the network for 35 epochs using the ADAM optimiser~\cite{kingma2014adam} while assigning 25 positive pairs and 25 negative pairs for each batch. The learning rate is initially set to 0.0001 and divided by a factor of 10 every 5 epochs. After training the network once, the weights are fixed for all experiments without any fine-tuning.

\subsection{Generating Grasp Proposals}

\begin{figure}[t!]
\begin{center}
\includegraphics[width=\textwidth]{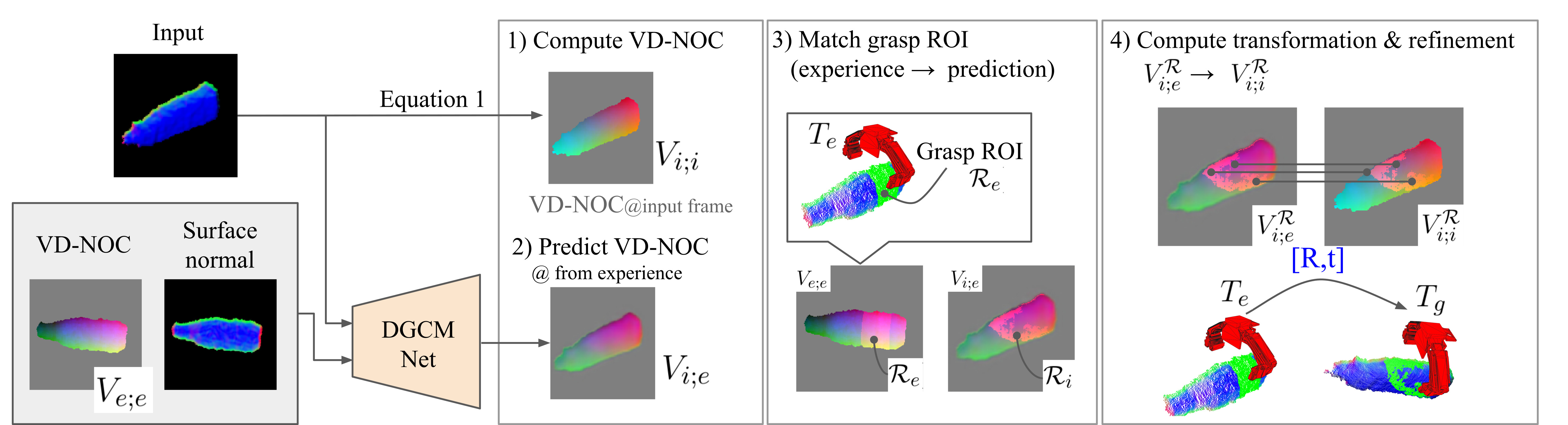}
\end{center}
\caption{Overview of the process for generating grasp proposals from the nearest neighbour experience.}\label{fig:grasp_proposal}
\end{figure}

The overview in Figure~\ref{fig:est_process} shows the process of retrieving and generating grasps given an input depth image. First, the surface normal image of the input is encoded to a feature map $f_i$ by the geometry encoder. This is compared to all feature maps $\{f_e\} \hspace{1ex} \forall \hspace{1ex} e \in \mathcal{E}$ to find a set of nearest neighbours $\mathcal{N}_i \subset \mathcal{E}$. The stored VD-NOC values $V_e$ of a sample $e \in \mathcal{N}_i$ is loaded to compute the VD-NOC feature map $c_e$. Given $c_e$, $f_e$ and $f_i$, the decoder predicts the VD-NOC values of the input depth image $V_{i;e}$ (VD-NOC values of $D_i$ in the frame of $D_e$) as shown in Figure~\ref{fig:grasp_proposal}. The ROI of the experience $\mathcal{R}_e$ is used to compute the corresponding ROI for the input $\mathcal{R}_i = \{p \in V_i : \underset{e}{\min}|p - p_e| < \theta_{c} \hspace{1ex} \forall \hspace{1ex} p_e \in \mathcal{R}_e\}$, which is the subset of points whose distance to the nearest points in $\mathcal{R}_e$ is below a threshold $\theta_{c}$. The predicted VD-NOC ROI points is denoted $V_{i;e}^\mathcal{R}$ and are defined in the camera frame of $D_e$. Each pixel of $V_{i}^\mathcal{R}$ forms a 3D-3D correspondence from the VD-NOC values $V_{i;e}^\mathcal{R}$ and $V_{i;i}^\mathcal{R}$ that are defined in $D_e$ and $D_i$. Thus, an initial rotation from the camera frame of the experience to the camera frame of the input is derived by aligning the ROI VD-NOC images. The grasp pose $T_e$ is then aligned to $D_i$ by computing the rotation that minimises the summation of distances of the correspondences given by,
\begin{equation}\label{eq:initial_align}
 R_\textrm{init},t_\textrm{init} = \underset{R,t}{\arg\min} \sum_{\mathcal{R}_i}{|| (RV_{i;e}^\mathcal{R} + t) - V_{i;i}^\mathcal{R} ||_{2}}.
\end{equation}

The optimised solution for Equation~\ref{eq:initial_align} is obtained using singular value decomposition. The unit of $t_\textrm{init}$ does not correspond to the scale of the 3D space because the VD-NOC values are normalised. Therefore, the translation $t_\textrm{init}$ is separately computed using the difference between the mean coordinates between the maximum and minimum values of the ROI points as was applied in Equation~\eqref{eq:vd-noc}. The computed rotation and translation are used to transform all ROI points of the experience $\mathcal{R}_e$ to the scene and the alignment is refined by applying the iterative closest point algorithm. The grasp pose in the experience $T_e$ is transformed to create the grasp proposal $T_g$ by applying the same refined transformation. Finally, the gripper position is moved a fixed distance from the object surface by translating along the approach direction with respect to the closest point in the input.

Each match in the database has an associated score in the range $(0,1]$ that represents that similarity of the depth image to the input, which is used as a pseudo-measure for the quality of the grasp. This score is computed as,
\begin{equation}\label{eq:score}
    S(i, e) = e^{-||f_i - f_e||_2}.
\end{equation}
The final output is a set of grasps $\mathcal{G} = \{(T_g, s_g)\}$ where each grasp proposal is composed of a transformation of the gripper into the scene $T_g$ as well as a score value $s_g$ using Equation~\eqref{eq:score}.


\section{Offline Experiments}\label{sec:ablation_studies}
This section analyses the grasp proposal method with a hand annotated dataset. Experiments are performed to first investigate the quality of grasp pose prediction with respect to the size of the grasp experience database and secondly to evaluate the ability to transfer grasps between observations of objects within the same and to different classes. The threshold for matching ROI correspondences $\theta_{c}$ is set to 0.3 for all experiments. This value produces reasonable separation of ROI areas and other parts of objects. Every
stored experience is duplicated with in-plane rotations at angles between -90 and +90 degrees with a step
size of 45 degrees. This enables grasp transfer to objects in new poses. Code for DGCM-Net is publicly available at \url{https://rgit.acin.tuwien.ac.at/v4r/dgcm-net}.

\subsection{Dataset}

\begin{figure}[t!]
\begin{center}
\includegraphics[width=\textwidth]{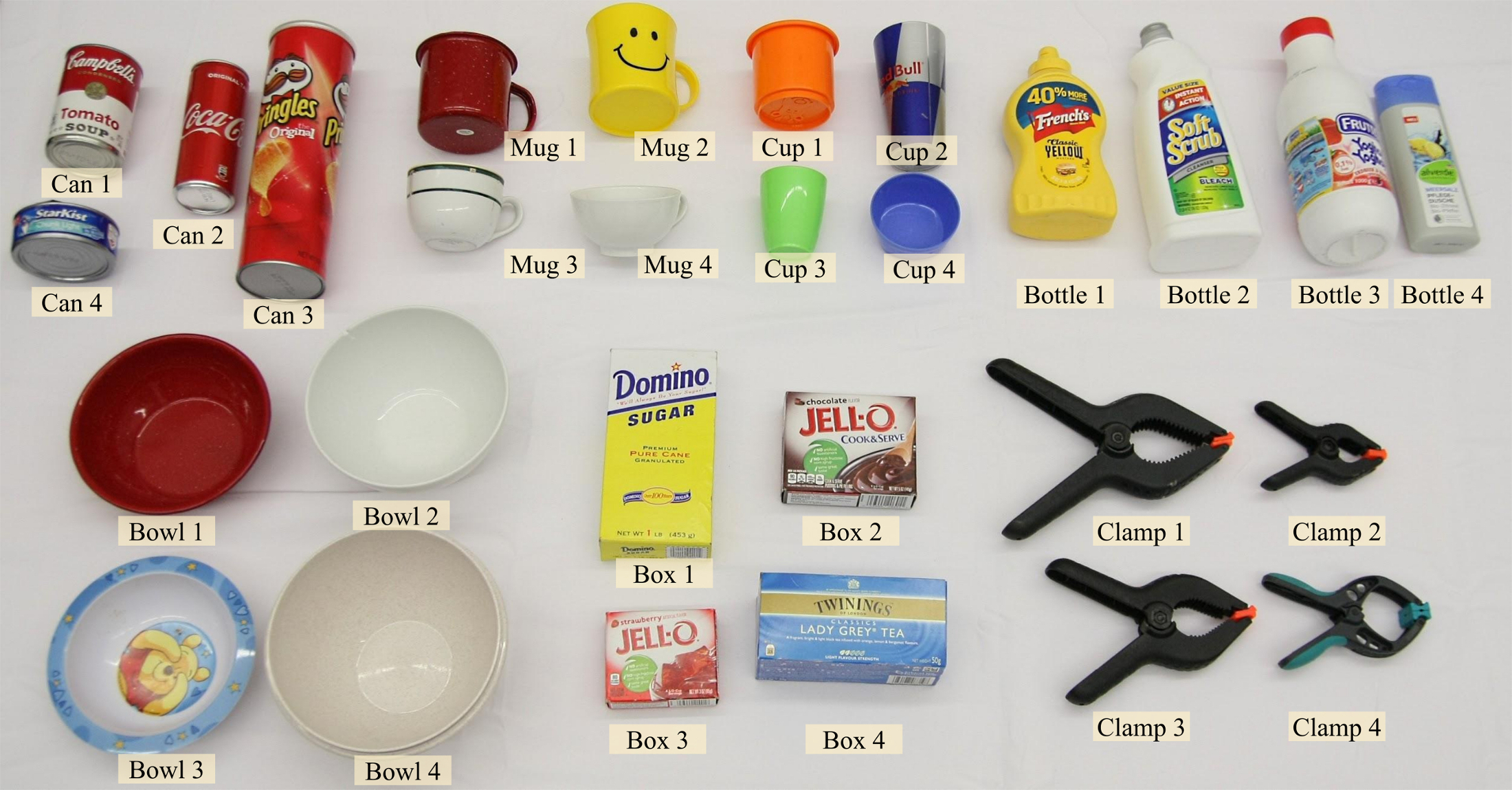}
\end{center}
\caption{Objects in the dataset used for the offline experiments and to supplement past experience during the online experiment.}\label{fig:object_dataset}
\end{figure}

A dataset is created to evaluate the quality of grasp prediction comprising depth images of the objects shown in Figure~\ref{fig:object_dataset}. These objects are organised into seven classes: \texttt{can}, \texttt{mug}, \texttt{cup}, \texttt{bottle}, \texttt{bowl}, \texttt{box} and \texttt{clamp}. Four instances are used for each class and a number of these instances are from the YCB object dataset~\cite{Calli2017}, while other instances are objects commonly found in homes. The dataset is available at \url{https://www.acin.tuwien.ac.at/en/vision-for-robotics/software-tools/lfed-6d-dataset/}.

Recordings are made by placing each object on a small table and capturing a depth image with an ASUS XTion Pro Live RGB-D camera. Each object is placed in various poses and locations, and the camera is moved between two different heights. The object is segmented in each depth image by detecting the table surface with RANSAC and selecting all points that remain above the table plane. The dataset does not require ground truth segmentation, but instead should be segmented by the same method that extracts the masks for the input images in order for the entries in the experience database to best resemble the inputs.

Grasp poses for a parallel-jaw gripper are manually annotated in the depth images. Each depth image consists of possibly multiple grasp annotations according to their direction, for example, from the top or from the side. The full dataset used for testing consists of depth images, segmentation masks and grasp poses for 28 objects of interest. 

\subsection{Measuring Grasp Pose Quality}
Reporting quantitative statistics requires the quality of the estimated grasp poses to be measured. It is possible to execute physics simulation and to check for grasp success, however, to isolate the grasp prediction itself, we measure the difference in grasp pose for an input with respect to the annotated pose. The experiments are simplified by selecting only grasp annotations on the top of the objects when they are placed in their upright canonical pose. Even though the grasp proposals are limited to top down, multiple poses are available, especially for objects that are symmetric or are elongated in the x or y dimension such that translations of a top-down grasp are equivalent. Consistency of top-down grasp poses is ensured by testing with the subset of classes \texttt{can}, \texttt{mug}, \texttt{cup} and \texttt{bottle}.

A grasp pose is regarded as correct when the translation error is less than 5cm and the rotational error around the x- and y-axes is less than 15 degrees. A rotation error around the z-axis in the gripper frame, which is parallel to the rotational axis of an object, is ignored since it should be a successful grasp regardless of the rotation with respect to this axis. 

\subsection{Increasing Experience}

\begin{figure}[t!]
\begin{center}
\includegraphics[width=\textwidth]{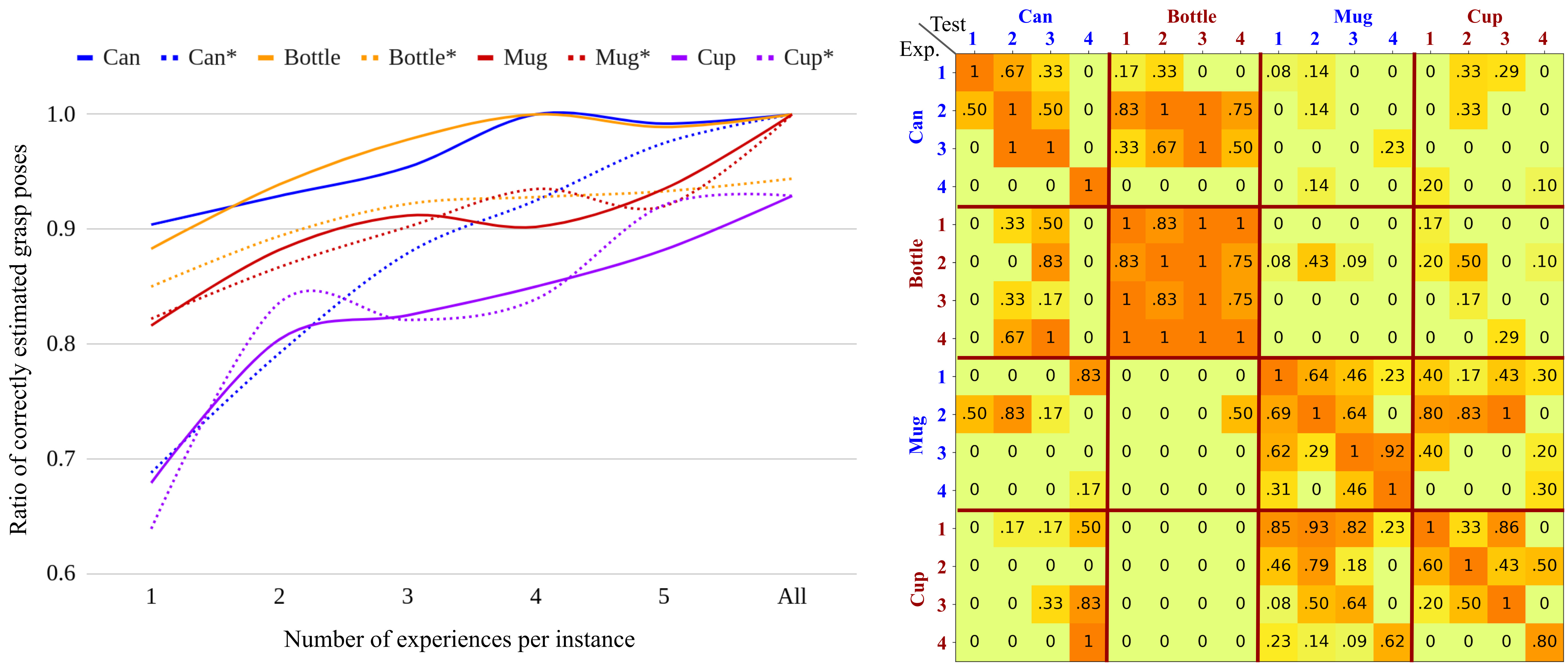}
\end{center}
\caption{Results of the offline experiments. Left: Ratio of accurately estimated grasp pose with increasing number of experiences per instance. Solid lines show results when class specific experience is used for each test class. Dotted lines show results when experience from all classes is used. Right: Ratio of accurately estimated grasp poses using experience from each instance in all classes.}\label{fig:ablation_result}
\end{figure}

The left plot of Figure~\ref{fig:ablation_result} shows the ratio of correctly estimated grasp poses with increasing number of experience per instance. Solid lines show results when only the experience for the relevant class is considered and the dotted lines show the results when experience for all classes is considered. For each configuration (class and number of experience per instance), we perform ten iterations using randomly selected samples in the iteration. The results with class specific experience demonstrate that the grasp poses are often correctly estimated even if only one experience is included per instance. For the \texttt{can} and \texttt{bottle} classes, the correct estimation is approximately 90\%, while the worst performing class, \texttt{cup}, achieves 68\%. However, as the number of experiences increases per instance in each class, the grasp pose estimation improves. 

The dotted lines show the variation in performance when including other classes for experience, which reflects more practical scenarios in the real world. Except for the \texttt{mug} class, the performance slightly drops because the retrieval of experiences from different classes can cause inaccurate prediction of VD-NOC values. However, the accuracy still achieves more than 79\% when two experiences are included per instance. This implies that the feature space encoded by the geometry encoder is sufficient to distinguish different geometrical shapes. The performance gap for the \texttt{can} class, which has the most simple shape, is comparably larger than for the other classes. This is because the mapping from more complex to simpler geometries produces inaccurate estimations by transferring detailed shapes into simpler geometries. We discuss more detail regarding this relationship between classes in the following section.

\begin{figure}[t!]
\begin{center}
\includegraphics[width=\textwidth]{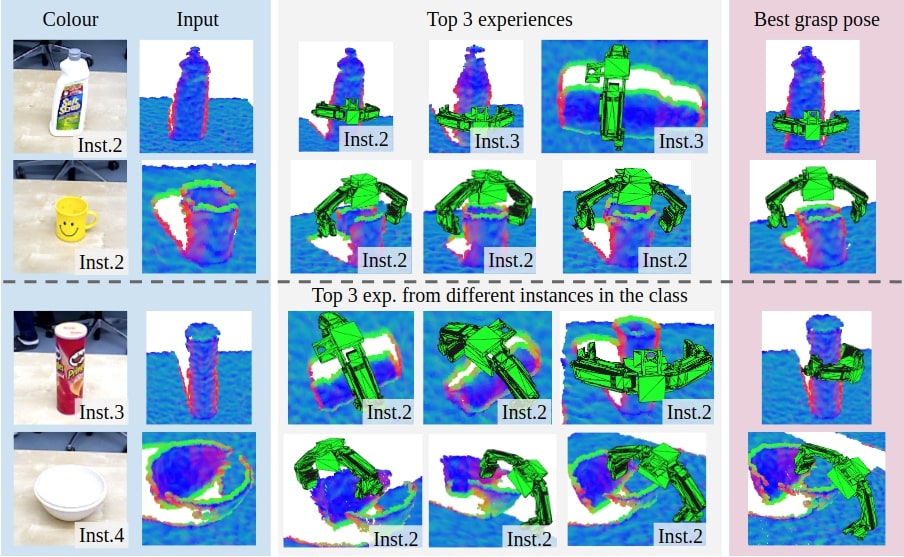}
\end{center}
\caption{Examples of the three nearest experiences and estimated grasp poses for example input images. Top two rows show examples when the same instance is included in the experience database. Bottom two rows show examples when the instance is not in the experience database.}\label{fig:evaluation_nn_results}
\end{figure}

Figure~\ref{fig:evaluation_nn_results} shows qualitative results. The figure shows the three nearest experiences and the best transformed pose for different instances from different classes. The first and second rows are obtained when the experience database contains samples from all classes and include the exact instance in the test image. The third and fourth rows are obtained after excluding the instance in the test image so that different instances in the same class must be retrieved to generate grasp proposals. The results reveal that the grasp poses are transformed to similar locations and directions even if object poses from experience are different (see the grasp poses for the \texttt{can} in the third row).

\subsection{Transfer Between Instances and Classes}
These experiments show that experience can be transferred between instances and classes. The experiments are conducted by using all experience from a single object instance while testing on different instances. The evaluation metric and subset of target grasp poses are the same as in the previous experiment. The matrix on the right of Figure~\ref{fig:ablation_result} shows that the experience of instances transfer well to other instances within the same class. Furthermore, experience also transfers beyond the class. For example, many good grasps are found for \texttt{bottle} instances when provided by experience from \texttt{can} instances (both types of objects have a closed surface on the top) and that instances of the \texttt{cup} class provide sufficient experience for grasping \texttt{mug} instances (both types of objects have no surface on the top). However, the results show that it is difficult to transfer the experience of a class to an instance in the same class when the geometry and scale of the instance are different from the other instances in the class (e.g., \texttt{Can4} and \texttt{Cup4}). Thus, better grasp poses are obtained when the experience is obtained from geometrically similar objects regardless of explicit classes. It is also observed that grasps for simpler geometries (e.g., from \texttt{can} to \texttt{bottle} and \texttt{cup} to \texttt{mug}) are more accurately transferred, while the other direction from complex geometry to simpler geometry is more difficult. This is because the network tries to predict VD-NOC values of detailed shapes of objects in the experience set, such as handles of \texttt{mug} instances, which potentially causes errors by predicting corresponding points even if the shapes are missing in the new object.


\section{Robot Experiments}\label{sec:experiments}

This section presents results of real-world grasping experiments with a mobile manipulator. First, we describe the hardware set up used for the experiments. Second, we compare our method to baseline approaches. Third, we evaluate the full pipeline of online incremental grasp learning. Finally, we demonstrate the extension to semantic grasp learning.

\subsection{Experimental Details}

The robot experiments are performed with the Toyota Human Support Robot~\cite{Yamamoto2019_HSR}. The platform consists of a 4-DOF arm but motions are computed including the omni-directional base, which effectively offers seven degrees of freedom. Motions for grasp execution are planned using MoveIt~\cite{Chitta2012_moveit}\footnote{http://moveit.ros.org}. The end-effector is a parallel-jaw gripper and grasp success is measured by checking the distance between the tips of the gripper after the target object is lifted. If the distance is non-zero, then the grasp is declared successful, otherwise, it is a failure. Depth images are captured with the onboard ASUS XTion Pro Live RGB-D sensor positioned on the head of the robot.

For all grasping experiments, individual objects are placed on a small table that has a height of 45cm. The robot is approximately positioned 30cm from the table (edge of robot base to edge of table). The head of the robot is tilted such that the camera faces the centre of the table. The torso of the robot is raised to give an approximate distance from the camera to an objects of 1m to suit the optimal range of the sensor. Objects are segmented from the table with the same procedure for generating segmentation masks for the dataset. 

All code is written in C++ and Python, and is running on the robot in Ubuntu 16.04. ROS~\cite{Quigley2009_ROS}\footnote{https://www.ros.org/} is used for process communication. DGCM-Net is implemented in Tensorflow and is running on an external PC with an NVIDIA GTX 1050~Ti. 

\subsection{Comparison to Baselines}\label{sec:comprison_baseline}

Experiments are conducted to measure the grasp performance of our framework. For comparison, experiments are also performed with a number of baselines. The full set of methods is as follows:

\begin{itemize}
    \item HAF: The approach introduced in~\cite{Fischinger2015_HAF}, where height accumulated features are extracted from point clouds to abstract grasp-relevant structure. The features are computed on different regions of the input and a support vector machine is trained to predict the quality of the grasp for each feature. Both top-down and forward-facing grasps are enabled, and the output with highest score is executed. We use the original code provided\footnote{https://github.com/davidfischinger/haf\_grasping}.
    \item GPD: The approach introduced in~\cite{tenPas2017_GPD}, where grasps are sampled using the surface geometry of the input point cloud. Grasp success for each sample is classified using a convolutional neural network (CNN). This takes as input three images: an averaged height map of occupied points, averaged height map of the unobserved region and averaged surface normals. Given this input, the CNN generates a score value. Finally, grasps are clustered and the highest scoring cluster is selected. We use the original code provided\footnote{https://github.com/atenpas/gpd} and the full 15 channel version.
    \item DGCM-Net: The grasp proposals from DGCM-Net using pre-collected experience for the relevant object classes in the experiments. Similar to GPD, the set of proposals from DGCM-Net are clustered and the highest scoring cluster is executed. Clustering is performed by grouping all grasps within 5cm translation and 15 degrees rotation. The grasp of the cluster is the mean pose of the proposals that make up the cluster. The cluster with the highest summed score is executed. The number of nearest neighbours to be retrieved by DGCM-Net is set to 10.
    \item GPD + DGCM-Net: Grasps are proposed using GPD and the scores are modified by the predictions from DGCM-Net. First, the grasps from the GPD method are computed and the scores are normalised to the range $[0, 1]$. Then DGCM-Net is run on the same input and for each GPD candidate, we find all DGCM-Net proposals within 5cm translation and 15 degrees rotation. The experience score is the average of the scores for all DGCM-Net grasps deemed to be nearby. The final score for each GPD candidate is the average of the normalised GPD score and the summed experience score. The grasp with the highest final score is executed.
\end{itemize}

Many robotic grasping approaches are successful for the bin-picking task, e.g.,~\cite{Mahler2017_DexNet2}. However, these are focused on 2D grasping and therefore expect a top-down view of the scene and only generate a grasp parallel to the camera axis. This are unsuitable for our robot platform due to the position of the arm on the front of the body that occludes the scene when facing the camera directly downwards. Additionally, bin-picking methods are at a disadvantage because they only generate grasps for a single approach direction. It is left to future work to extend the evaluation to this type of scenario.

\begin{figure}[t!]
\begin{center}
\includegraphics[width=\textwidth]{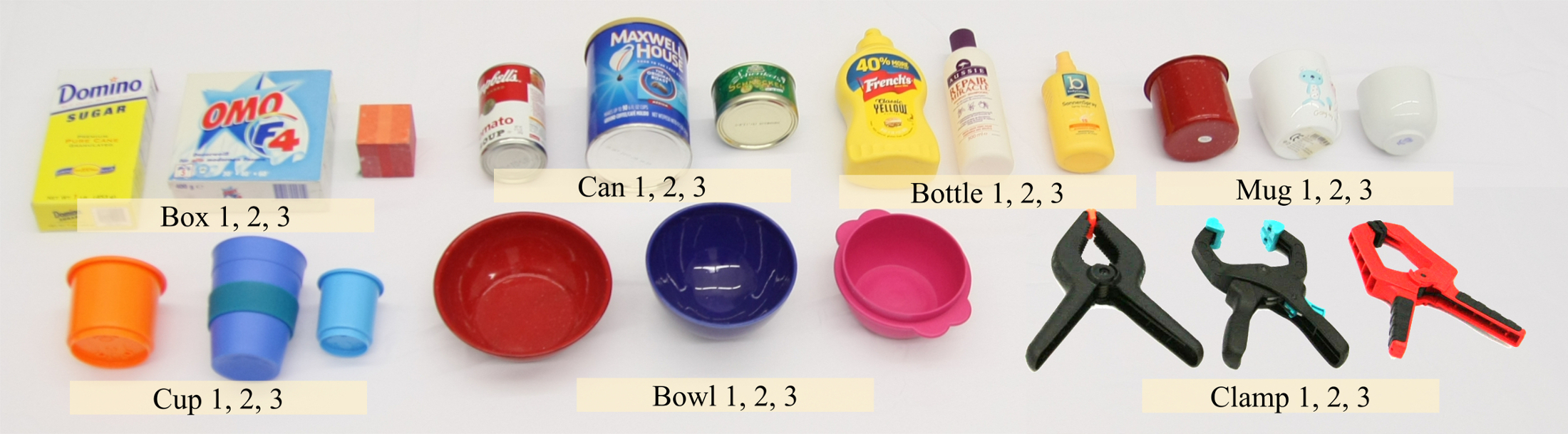}
\end{center}
\caption{Test objects used for the online grasping experiments. Three instances in seven classes are used.}
\label{fig:fig:online_experiment_objects}
\end{figure}

The experiments are performed for objects from the classes \texttt{box}, \texttt{can}, \texttt{bottle}, \texttt{mug}, \texttt{cup}, \texttt{bowl} and \texttt{clamp}. Three instances are chosen per class and five poses are considered per instance. The five poses for each instance are kept constant for the experiments with each grasping method. The objects selected for the experiments are shown in Figure~\ref{fig:fig:online_experiment_objects}. These include one object from the YCB set for each class from the objects used in Section~\ref{sec:ablation_studies}, in particular, the \texttt{sugar box}, \texttt{spam can}, \texttt{mustard bottle}, \texttt{red mug}, \texttt{orange cup}, \texttt{red bowl} and \texttt{XL clamp}. The other two objects for each class are a mixture of YCB objects and common objects found in homes.

The experience used for our method is an extension of the database from Section~\ref{sec:ablation_studies} that includes instances from the additional classes of \texttt{box}, \texttt{bowl} and \texttt{clamp}. Since we are interested in observing the grasp performance for unseen objects, the YCB objects selected as target objects are removed from the experience database.

\begingroup
\renewcommand{\arraystretch}{1.4}
\begin{table}[t!]
\begin{centering}
\begin{tabular}{p{2.7cm}>{\centering\arraybackslash}m{2.7cm}>{\centering\arraybackslash}m{2.7cm}>{\centering\arraybackslash}m{2.7cm}>{\centering\arraybackslash}m{2.7cm}}
 \toprule
  & HAF & GPD & DGCM-Net & GPD + DGCM-Net \\
  \hline
 Box & 0.87 & 0.80 & 0.67 & 1.00 \\ 
 Can & 0.87 & 0.67 & 0.93 & 0.73 \\
 Bottle & 0.87 & 0.93 & 0.93 & 0.93 \\
 Mug & 0.80 & 0.80 & 0.87 & 1.00 \\
 Cup & 0.80 & 0.80 & 0.73 & 1.00 \\
 Bowl & 0.40 & 0.87 & 0.80 & 0.87 \\
 Clamp & 0.40 & 0.60 & 0.60 & 0.67 \\
 \hline
 Average & 0.71 & 0.79 & 0.79 & 0.89 \\
 \bottomrule
\end{tabular}
\caption{Grasp success rate of our framework and baseline methods for different target object classes. The bottom row shows the average for all classes.}
\label{tab:grasping_comparison}
\end{centering}
\end{table}
\endgroup

Performance is measured by grasp success rate, which is the number of successful grasps divided by the total number of attempts. Table~\ref{tab:grasping_comparison} reports the average grasp success rate for each class and the average for all classes (bottom row). The results show that our method performs equivalently to GPD and that both methods outperform HAF (+8\%). However, combining experience with GPD achieves a much higher grasp success rate overall. In comparison to the original GPD method, this is an increase of 10\%. 

For most classes, the combined approach performs either the same as the best performing individual method or better. The only exception is the \texttt{can} class, which has 20\% lower grasp success rate than our direct method. Our observation during the experiments is that GPD often proposed grasps on and orthogonal to the rim of the \texttt{can} objects, which resulted in failures. This exposes the flaw that if the initial candidates are unfavourable, the combined approach cannot improve. For the cans, when grasp experience on the top of the can are stored, the grasps on the rim are still similar in position and orientation to warrant their selection.

Surprisingly, the \texttt{box} class is the most difficult for our approach despite having easy geometry to compute a grasp as shown by the high success rate of HAF. This can be explained by the fact that all boxes can be represented by a single \texttt{box} by changing the scale in the different dimensions. Thus, the network has to decide whether to regard a new box as a scaled version of an experience or as a transformed (i.e., rotated) box. The ambiguity causes noisy predictions of VD-NOC values. Furthermore, since grasp proposals are transformed from previous experiences and are ideally in a similar grasp location, a scale change may cause the prediction to exceed the range of the gripper, resulting in its rejection due to the collision.

\subsection{Incremental Learning}

This set of experiments demonstrate the full incremental learning framework. The test objects chosen are the \texttt{grey clamp} from our dataset, the \texttt{plastic drill} from the YCB object dataset and a \texttt{gaming controller}. Past experience is stored in the database, however, not for the classes of the test objects. Therefore, experience for \texttt{clamp} class is removed. Since good grasps may not be generated for the unseen objects, GPD is used in the beginning until DGCM-Net makes reasonable predictions. A threshold of 0.9 is set as the minimum feature distance that must be achieved by the output of DGCM-Net, otherwise, the best grasp from GPD is executed. Typically it only takes one or two successful attempts for the system to switch from GPD to DGCM-Net. The objects are placed randomly on the table at the beginning of each experiment and the system runs autonomously, with the robot grasping the object from where it lies after a successful or failed attempt. The object is only handled by a person if it is unintentionally moved near the edge of the table and presents a risk of falling. After successful grasps, objects are placed on the table by the robot and receive a slight variation in pose; failed grasps typically cause considerable object movement. After any grasp attempt, the robot base returns to the start position and the localisation inaccuracy generates
further viewpoint variation.

\begin{figure}[t!]
\begin{center}
\includegraphics[width=\textwidth]{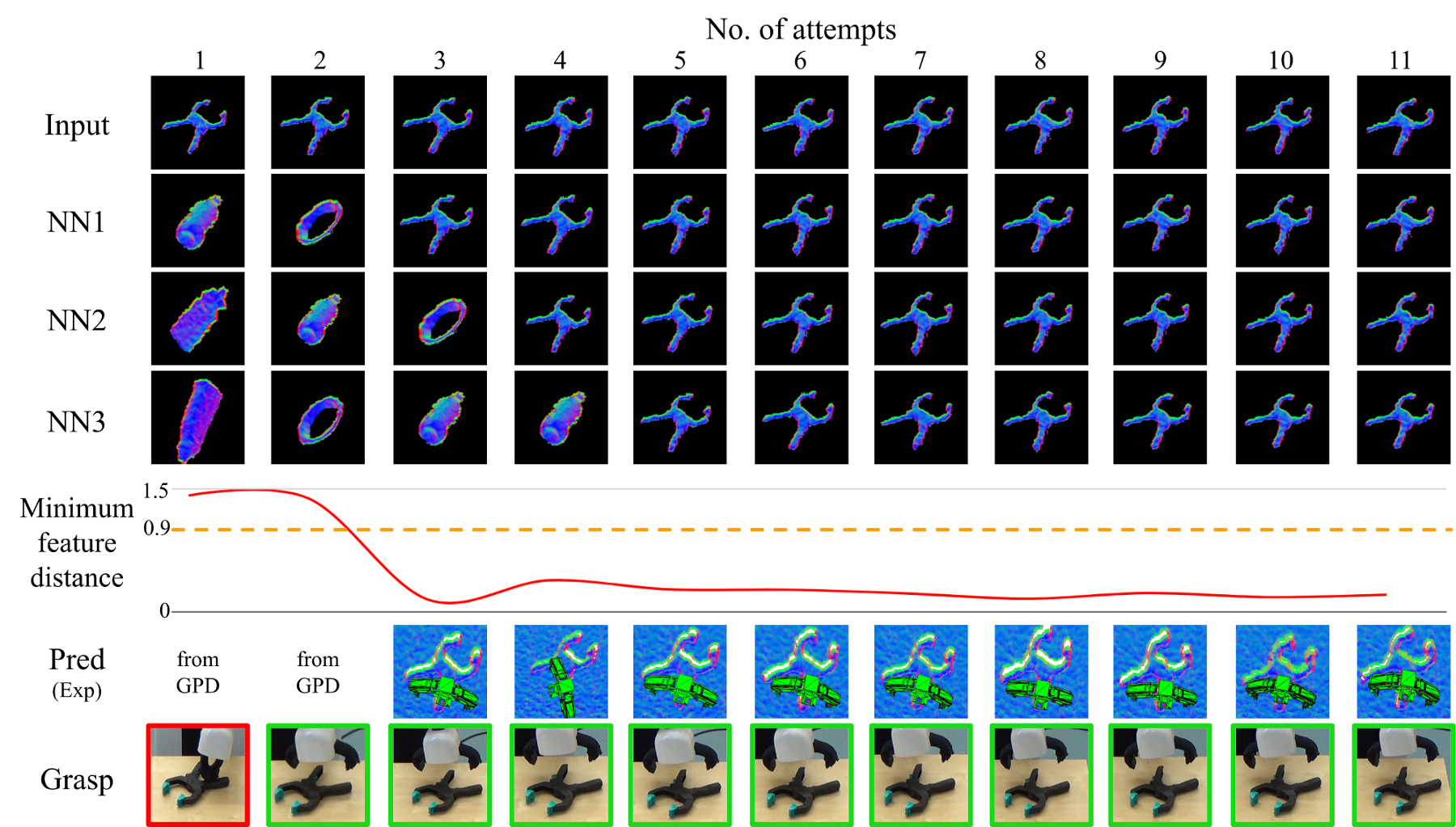}
\end{center}
\caption{Summary of results for the incremental learning experiment with the \texttt{grey clamp}.}
\label{fig:results_incremental_learning_clamp} 
\end{figure}

\begin{figure}[t!]
\begin{center}
\includegraphics[width=\textwidth]{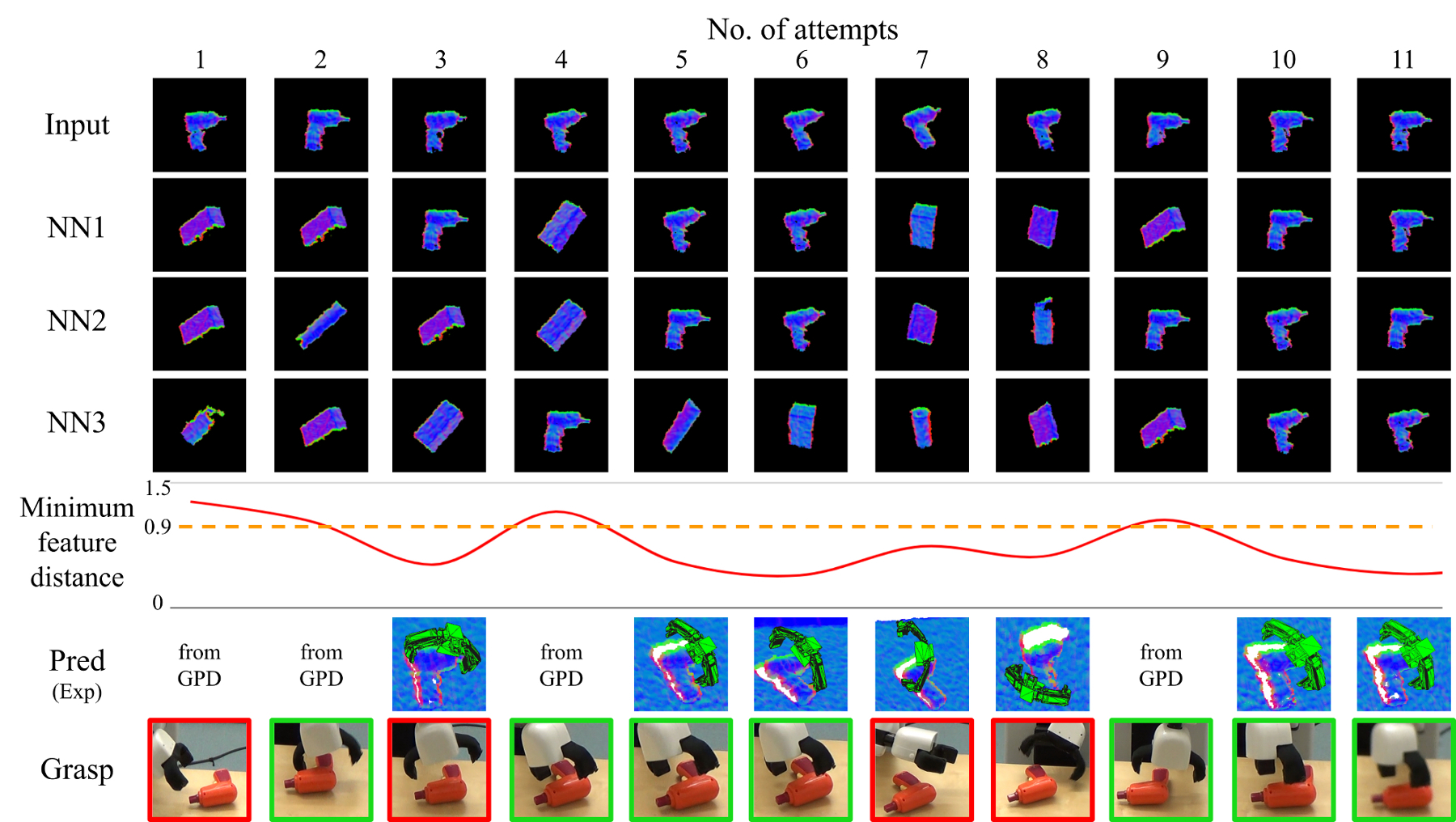}
\end{center}
\caption{Summary of results for the incremental learning experiment with the YCB \texttt{plastic drill}.}
\label{fig:results_incremental_learning_drill} 
\end{figure}

\begin{figure}[t!]
\begin{center}
\includegraphics[width=0.95\textwidth]{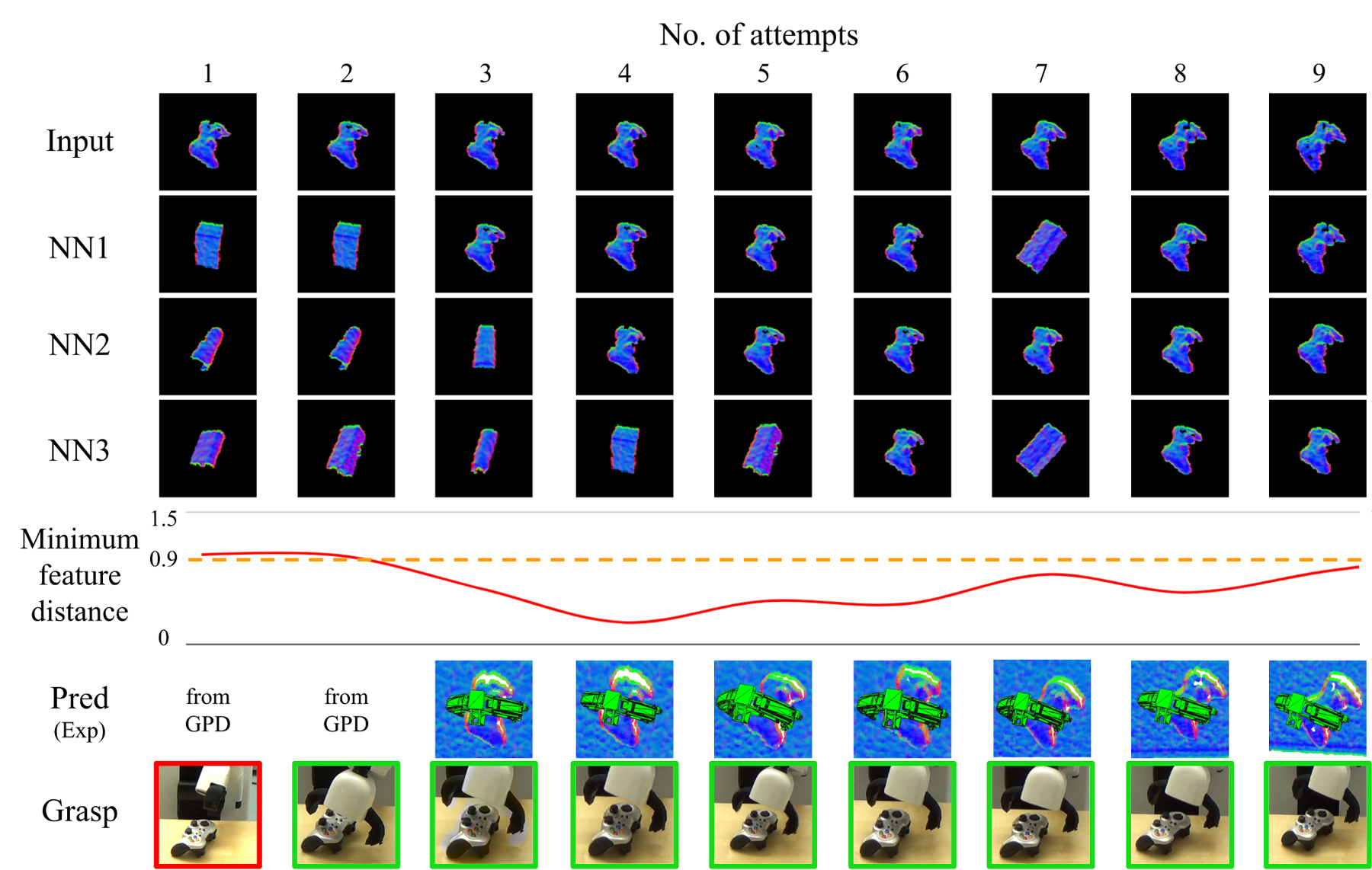}
\end{center}
\caption{Summary of results for the incremental learning with the \texttt{gaming controller}.}
\label{fig:results_incremental_learning_controller} 
\end{figure}

Figures~\ref{fig:results_incremental_learning_clamp},~\ref{fig:results_incremental_learning_drill} and~\ref{fig:results_incremental_learning_controller} show the evolution of grasp success for the three objects. In these figures, the first row shows the surface normal image of the input and the second to fourth rows show the nearest three matches in the database. Below this we plot the minimum feature distance of the nearest neighbour. Lastly, we show the best grasp proposal from DGCM-Net and the actual gripper position during the grasp captured by an external camera (red border indicates failure and green border indicates success). A video of the experiments is available at \url{https://youtu.be/iI_P1UVXfjo}.

From these experiments, we make two observations. Firstly, after the first successful attempt from GPD is recorded, the robot typically continues to grasp the target objects successfully. The predicted grasps for each attempt confirms that our method reliably predicts the same successful experience so long as the object and its shape is correctly identified. The second observation is that the minimum feature distance drops below the threshold after as many as one sample is in the database. This is most apparent for the \texttt{grey clamp} in which the minimum feature distance is very small for all subsequent trials.

The grasping for the \texttt{plastic drill} and \texttt{gaming controller} are less reliable than for the \texttt{grey clamp}. For the \texttt{plastic drill}, the system still exhibits some failures even after accumulating experience. In both cases, the nearest feature distance does not converge to the same low value as was observed for the \texttt{grey clamp}. The reason is that the \texttt{plastic drill} and \texttt{gaming controller} have less distinct shape and therefore are more difficult to match. This is especially noticeable when the objects have rotated. The objects are often confused as an instance from the \texttt{box} class and the grasps for the matching box is executed. Fortunately for the \texttt{gaming controller}, the execution still results in success. However, for the \texttt{plastic drill}, the predicted grasp is not very good and the grasp fails.

\subsection{Semantic Grasping}

A final set of experiments demonstrate the extension of our method to generate semantic grasps for instances belonging to the same functional class. For these experiments, we investigate grasps on the handles of \texttt{mug} and \texttt{drill} objects. The experience is hand annotated for test exemplars of instances similar to the target object. It is possible for the robot to self-learn semantic grasping by incorporating, for example, affordance detection to indicate whether the location of the grasp matches the affordance of the object part~\cite{Do2018_affordancenet}. This is out of scope for the present article and left for future work.

\begin{figure}[t!]
\begin{center}
\includegraphics[width=0.95\textwidth]{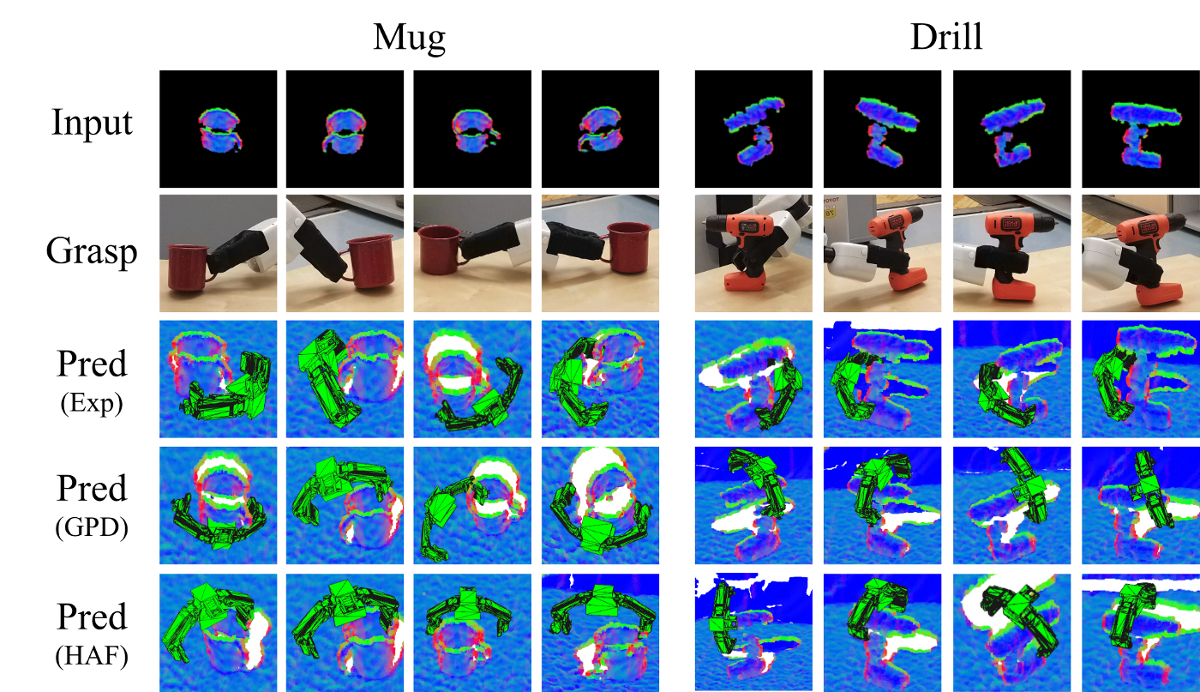}
\end{center}
\caption{Examples of semantic grasping for the YCB red \texttt{mug} and the YCB orange \texttt{drill} in four different poses. First row: Surface normal image. Second row: Grasp pose executed by the robot. Third row: Predicted grasp from DGCM-Net. Fourth row: Predicted grasp from GPD. Fifth row: Predicted grasp from HAF.}
\label{fig:results_semantic_grasp_examples} 
\end{figure}

Example grasp proposals generated by our method as well as GPD and HAF for a \texttt{mug} and a \texttt{drill} in various poses are shown in Figure~\ref{fig:results_semantic_grasp_examples}. Our method reliably generates grasp poses on the relevant object part, while both GPD and HAF fail to do so. Although the grasps from GPD and HAF may result in success, they do not support the functional use of the object. For the \texttt{mug}, it is understandable that the handle is not grasped because the quality of the depth data on that part of the object is very poor and does not characterise a stable grasp. Our method, on the other hand, does not only rely on the local structure to estimate the grasp. So long as there is some cue about the handle, as is present in these selected examples, the handle grasp is generated. The \texttt{drill} offers more depth data on the handle but the baselines still prefer to grasp the head. HAF is executed to find both top and front grasps, and the best scoring grasp is shown. Nonetheless, a top grasp or a front grasp on the head is preferred instead of the handle. A video of the grasps executed by the robot is available at \url{https://youtu.be/iI_P1UVXfjo}.


\section{Conclusion}\label{sec:conclusion}

This article presented an approach for incrementally learning grasps by leveraging past experience. In our system, every successful grasp is stored in a database and retrieved to guide future grasps. This is accomplished with the dense geometric correspondence network that is trained to predict the similarity between newly acquired input depth images and stored experiences as well as to predict 3D-3D correspondences to transform grasp poses. A descriptive feature space is constructed for the retrieval task using metric learning and correspondences are established by predicting view-dependent normalised object coordinate values.

Offline studies with a dataset showed that our approach precisely recovers grasps from experiences with the same object and also transfers well to unseen objects from the same or different class. Furthermore, results showed that more experience leads to more reliable grasp proposals. Hardware experiments with a mobile manipulator showed that our experience-based grasping method performs equally successful as the baseline and integration with the baseline shows overall superior performance. Additional experiments demonstrated the full online capability to efficiently learn grasps for unseen objects, often needing only one or two successful grasps to reliably re-grasp the same object. Finally, an extension was demonstrated whereby specific grasps, such as those on handles, can be desired in order to achieve semantically meaningful grasps.

One direction for future work is to include more instances per class when training DGCM-Net to better generalise over varied shapes of a class instead of simply manipulating scales of an object for each class. Furthermore, it has been observed that points on the bottom of objects are missing from the object masks since the points are regarded as table or background. Thus, more detail about objects would be extracted by applying a segmentation method that refines the mask using colour information. Another avenue of future work is to include failed experience during the learning phase. Particularly for objects that are difficult to grasp, it may take a large number of attempts to finally succeed. By also considering failures, it would be possible to reject grasp candidates and thus more quickly guide grasping to successful regions. Currently, our method was only tested with a parallel-jaw gripper. It would be interesting to extend this work to other hardware, such as three-finger grippers or anthropomorphic hands. It would be even more interesting to investigate how to transfer grasps between grippers so that experience learned by one platform can be exploited by another platform with different hardware.


\bibliographystyle{plain}
\bibliography{dgcm-net-patten-park-vincze-2020}

\end{document}